\documentclass[a4paper,twoside]{article}

\usepackage{epsfig}
\usepackage{subcaption}
\usepackage{calc}
\usepackage{amssymb}
\usepackage{amstext}
\usepackage{amsmath}
\usepackage{amsthm}
\usepackage{multicol}
\usepackage{pslatex}
\usepackage{apalike}

\usepackage{graphicx}
\usepackage{multirow}
\usepackage{array}
\usepackage{csquotes}
\newcommand{\dq}[1]{\enquote{#1}}

\newcommand{\us}{\rule{.4em}{.4pt}}
\usepackage{hyperref}
\usepackage{url}

\usepackage{SCITEPRESS}  

\begin{document}

\title{Sentiment Analysis of Czech Texts: An Algorithmic Survey}

\author{\authorname{Erion \c{C}ano\sup{1}, Ond{\v r}ej Bojar\sup{1}}
\affiliation{\sup{1}Institute of Formal and Applied Linguistics, Charles University, Prague, Czech Republic}
\email{\{cano, bojar\}@ufal.mff.cuni.cz}
}


\abstract{In the area of online communication, commerce and transactions, analyzing sentiment polarity of texts written in various natural languages has become crucial. While there have been a lot of contributions in resources and studies for the English language, \dq{smaller} languages like Czech have not received much attention. In this survey, we explore the effectiveness of many existing machine learning algorithms for sentiment analysis of Czech Facebook posts and product reviews. We report the sets of optimal parameter values for each algorithm and the scores in both datasets. We finally observe that support vector machines are the best classifier and efforts to increase performance even more with bagging, boosting or voting ensemble schemes fail to do so.}

\onecolumn \maketitle \normalsize \vfill

\section{INTRODUCTION}
\label{sec:intro}
%
Sentiment Analysis is considered as the automated analysis of sentiments, emotions or opinions expressed in texts towards certain entities \cite{citeulike:14184846}. The proliferation of online commerce and customer feedback has significantly motivated companies to invest in intelligent text analysis tools and technologies where sentiment analysis plays a crucial role. There have traditionally been two main approaches to sentiment analysis. The first one uses unsupervised algorithms, sentiment lexicons and word similarity measures to \dq{mine} emotions in raw texts. The second uses emotionally-labeled text datasets to train supervised (or deep supervised) algorithms and use them to predict emotions in other documents. 
%
Naturally, most of sentiment analysis research has been conducted for the English language. Chinese \cite{ZHANG2018395,peng2017review,Wu2015} and Spanish \cite{TELLEZ2017457,MIRANDA2017} have also received a considerable extra attention in the last years. \dq{Smaller} languages like Czech have seen fewer efforts in this aspect. It is thus much easier to find online data resources for English than for other languages \cite{7325106}.
%
One of the first attempts to create sentiment annotated resources of Czech texts dates back in 2012 \cite{DBLP:conf/konvens/VeselovskaHS12}. Authors released three datasets of news articles, movie reviews, and product reviews. A subsequent work consisted in creating a Czech dataset of information technology product reviews, their aspects and customers' attitudes towards those aspects \cite{biblio-3441598209493476448}. This latter dataset is an essential basis for performing aspect-based sentiment analysis experiments \cite{S16-1059}. Another available resource is a dataset of ten thousand Czech Facebook posts and the corresponding emotional labels \cite{W13-1609}. The authors report various experimental results with Support Vector Machine (SVM) and Maximum Entropy (ME) classifiers.     
%
Despite the creation of the resources mentioned above and the results reported by the corresponding authors, there is still little evidence about the performance of various techniques and algorithms on sentiment analysis of Czech texts. 
%
In this paper, we perform an empirical survey, probing many popular supervised learning algorithms on sentiment prediction of Czech Facebook posts and product reviews. We perform document-level analysis considering the text part (that is usually short) as a single document and explore various parameters of \emph{Tf-Idf} vectorizer and each classification algorithms reporting the optimal ones. According to our results, SVM (Support Vector Machine) is the best player, shortly followed by Logistic Regression (LR) and Na\"ive Bayes (NB). Moreover, we observe that ensemble techniques like Random Forests (RF), Adaptive Boosting (AdaBoost) or voting schemes do not increase the performance of the basic classifiers. 
%
The rest of the paper is structured as follows: Section~\ref{sec:datasets} presents some details and statistics about the two Czech datasets we used. Section~\ref{sec:prepvect} describes the text preprocessing steps and vectorizer parameters we grid-searched. Section~\ref{sec:alg} presents in details the grid-searched parameters and values of all classifiers. In Section~\ref{sec:results}, we report the optimal parameter values and test scores in each dataset. Finally, Section~\ref{sec:disc} concludes and presents possible future contributions.  
\section{DATASETS}
\label{sec:datasets}
%
\begin{table}[!t]
\centering
\begin{tabular}{ | l r r | }
\hline
\bf Attribute & \bf Mall & \bf Facebook \\ [0.1ex] 
\hline
Records & 11K & 10K \\ [0.07ex]
Tokens & 151K & 105K \\ [0.07ex]
Av. Length & 13 & 10 \\ [0.07ex]
Classes & 2 & 3 \\ [0.07ex]
Negative & 4356 & 1991 \\ [0.07ex]
Neutral & - & 5174 \\ [0.07ex]
Positive & 7274 & 2587 \\ [0.07ex]
\hline
\end{tabular}
\caption{Statistics about the two datasets}
\label{table:dataStats}
\end{table}
\subsection{Czech Facebook Dataset}
\label{ssec:fb}
\begin{figure}[!h]
	\centering
	{\epsfig{file = 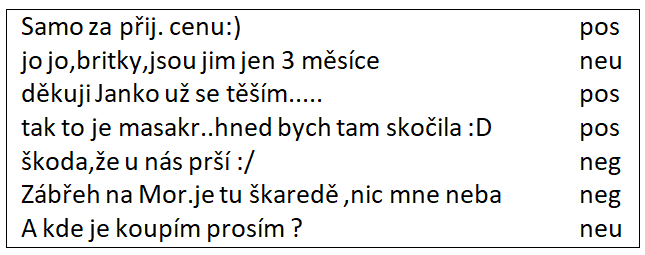, width = 7.4cm}}
	\caption{Samples from Czech Facebook dataset}
	\label{fig:fbData}
\end{figure}

%
Czech Facebook dataset was created by collecting posts from popular Facebook pages in Czech \cite{W13-1609}. The ten thousand records were independently revised by two annotators. Two other annotators were involved in cases of disagreement. To estimate inter-annotator agreement, they used Cohen's kappa coefficient which was about 0.66. Each post was labeled as \emph{negative}, \emph{neutral} or \emph{positive}. There were yet a few samples that revealed both \emph{negative} and \emph{positive} sentiments and were marked as \emph{bipolar}. Same as the authors in their paper, we removed the \emph{bipolar} category from our experimental set to avoid ambiguity and used the remaining 9752 samples. A few data samples are illustrated in Figure~\ref{fig:fbData}.    
\subsection{Mall.cz Reviews Dataset}
\label{ssec:mall}
\begin{figure}[!h]
	\centering
	{\epsfig{file = 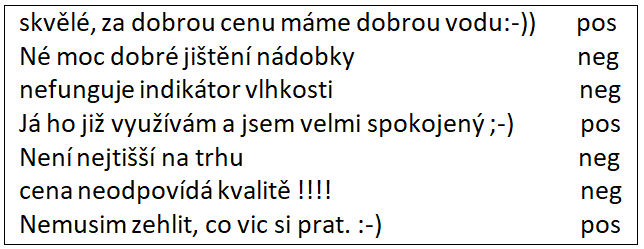, width = 7.4cm}}
	\caption{Samples from Mall reviews dataset}
	\label{fig:mallData}
\end{figure}
%
The second dataset we use contains user reviews about household devices purchased at \url{mall.cz} \cite{DBLP:conf/konvens/VeselovskaHS12}. The reviews are evaluative in nature (users apprising items they bought) and were categorized as \emph{negative} or \emph{positive} only. Some minor problems they carry are the grammatical or typing errors that frequently appear in their texts \cite{veselovska-2017}. In Table~\ref{table:dataStats} we present some rounded statistics about the two datasets. As we can see, Mall product reviews are slightly longer (13 vs. 10 tokens) than Czech Facebook posts. We also see that the number of data samples in each sentiment category are unbalanced in both cases. A few samples of Mall reviews are illustrated in Figure~\ref{fig:mallData}.
%
\begin{table}[!t]
\centering
\newcolumntype{C}[1]{>{\centering\arraybackslash} m{#1}}
\begin{tabular}{ | C{1.4cm} C{1.67cm} C{2.32cm} | }
\hline
\bf Vectorizer & \bf Parameters & \bf GS Values \\ 
\hline
\multirow{4}{*}{Tf-Idf} & ngram\us range & (1,1),~~(1,2),~~(1,3) \\ 
& stop\us words & Czech,~~None \\ 
& smooth\us idf & True,~~False \\ 
& norm & l1,~~l2,~~None \\
\hline
\end{tabular}
\caption{Tf-Idf vectorizer grid-searched parameters}
\label{table:vectpars}
\end{table}
\section{PREPROCESSING AND VECTORIZATION}
\label{sec:prepvect}
%
Basic preprocessing steps were applied to each text field of the records. First, any remaining markup tags were removed, and everything was lowercased. At this point, we saved all smiley patterns (e.g., \emph{\dq{:P}}, \emph{\dq{:)}}, \emph{\dq{:(}}, \emph{\dq{:-(}}, \emph{\dq{:-)}}, \emph{\dq{:D}}) appearing in each record. Smileys are essential features in sentiment analysis tasks and should not be lost from the further text cleaning steps. Stanford CoreNLP\footnote{\url{https://nlp.stanford.edu/software/tokenizer.shtml}} tokenizer was employed for tokenizing. Numbers, punctuation, and special symbols were removed. At this point, we copied back the smiley patterns to each of the text samples. No stemming or lemmatization was applied.  
\par
As vectorizer, we chose to experiment with \emph{Tf-Idf} which has been proved very effective with texts since long time ago \cite{10.1007/BFb0026683,1174522}. \emph{Tf-Idf} gives the opportunity to work with various n-grams as features (\emph{ngram\us range} parameter). We limited our experiments to single words, bigrams, and trigrams only since texts are usually short in both datasets. It is also very common in such experiments to remove a subset of words known as stop words that carry little or no semantic value. In our experiments we tried with full vocabulary or removing Czech stopwords that are defined at \url{https://pypi.org/project/stop-words/} package. Other parameters we explored are \emph{smooth\us idf} and \emph{norm}. The former adds one to document frequencies to smooth \emph{Idf} weights when computing \emph{Tf-Idf} score. The latter is used to normalize term vectors (None for no normalization). The parameters and the corresponding grid-searched values of \emph{Tf-Idf} are listed in Table~\ref{table:vectpars}.
%
\begin{table*}[!t]
\centering
\newcolumntype{C}[1]{>{\centering\arraybackslash} m{#1}}
\begin{tabular}{ | C{1.5cm} C{2.4cm} C{6.9cm} | }
\hline
\vspace{1mm}
\bf Algorithm & \bf Parameters & \bf Grid-Searched Values \\ 
\hline
\multirow{3}{*}{SVM} 
& C & 0.0001, 0.001, 0.01, 0.1, 1, 10, 100, 1000 \\ 
& gamma & 0.5, 0.1, 0.05, 0.01, 0.005, 0.001, 0.0005, 0.0001 \\ 
& kernel & linear, rbf, poly, sigmoid \\
\hline
\multirow{2}{*}{NuSVM} 
& nu & 0.35, 0.4, 0.45, 0.5, 0.55, 0.6, 0.65 \\ 
& kernel & linear, rbf, poly, sigmoid \\
\hline
\multirow{3}{*}{RF} 
& max\us depth & None, 10, 20, 30, 40, 50, 60, 70, 80, 90 \\ 
& max\us feat & 10, 20, 30, 40, 50, sqrt, None \\
& n\us est & 50, 100, 200, 400, 700, 1000 \\ 
\hline
\multirow{3}{*}{LR} 
& C & 0.0001, 0.001, 0.01, 0.1, 1, 10, 100, 1000 \\ 
& class\us weight & balanced, None \\
& penalty & l1, l2 \\ 
\hline
\multirow{4}{*}{MLP} 
& alpha & 0.0001, 0.0005, 0.001, 0.005, 0.01, 0.05, 0.1, 0.5 \\ 
& layer\us sizes & $ (10, 20, 40, 60, 80, 100) \times (1, 2, 3, 4) $ \\
& activation & identity, logistic, tanh, relu \\ 
& solver & lbfgs, sgd, adam \\ 
\hline
\multirow{2}{*}{NB} 
& alpha & 0.0001, 0.0005, 0.001, 0.005, 0.01, 0.05, 0.1, 0.5\\ 
& fit\us prior & True, False \\ 
\hline
\multirow{1}{*}{ME} 
& method & gis, iis, megam, tadm \\
\hline
\end{tabular}
\caption{Grid-searched parameters and values of each algorithm}
\label{table:algpars}
\end{table*}
\par
Besides using this traditional approach based on \emph{Tf-Idf} or similar vectorizers, it is also possible to analyze text by means of the more recent dense representations called word embeddings \cite{Mikolov:2013:DRW:2999792.2999959,pennington2014glove}. These embeddings are basically dense vectors (e.g., 300 dimensions each) that are obtained for every vocabulary word of a language when large text collections are fed to neural networks. The advantage of word embeddings over bag-of-word representation and \emph{Tf-Idf} vectorizer is their lower dimensionality which is essential when working with neural networks. It still takes a lot of text data (e.g., many thousands of samples) to generate high-quality embeddings and achieve reasonable classification performance \cite{10.1007/978-3-319-59569-6_42}. A neural network architecture for sentiment analysis based on word embeddings is described by \cite{Cano:2018:DLA:3220228.3220229}. We applied that architecture on the two Czech datasets we are using here and observed that there was severe over-fitting, even with dropout regularization. For this reason, in the next section, we report results of simpler supervised algorithms and multilayer perceptron only, omitting experiments with deeper neural networks.  
\section{SUPERVISED ALGORITHMS}
\label{sec:alg}
%
We explored various supervised algorithms that have become popular in recent years and grid-searched their main parameters. Support Vector Machines have been successfully used for solving both classification and regression problems since back in the nineties when they were invented \cite{Boser:1992:TAO:130385.130401,cortes1995support}. They introduced the notion of hard and soft margins (separation hyperplanes) for optimal separation of class samples. Moreover, the \emph{kernel} parameter enables them to perform well even with data that are not linearly separable by transforming the feature space \cite{Kocsor:2004:AKF:1008633.1008643}. The \emph{C} parameter is the error penalty term that tries to balance between a small margin with fewer classification errors and larger margin with more errors. The last parameter we tried is \emph{gamma} that represents the kernel coefficient for \dq{rfb}, \dq{poly} and \dq{sigmoid} (non linear) kernels. 
%
The other algorithm we tried is NuSVM which is very similar to SVM. The only difference is that a new parameter (\emph{nu}) is utilized to control the number of support vectors.
%
Random Forests (RF) were also invented in the 90s \cite{Ho:1995:RDF:844379.844681,709601}. They average results of multiple decision trees (bagging) aiming for lower variance. Among the many parameters, we explored \emph{max\us depth} which limits the depth of decision trees. We also grid-searched \emph{max\us feat}, the maximal number of features to consider for best tree split. If \dq{sqrt} is given, it will use the square root of total features. If \dq{None} is given then it will use all features. Finally, \emph{n\us est} dictates the number of trees (estimators) that will be used. Obviously, more trees may produce better results but they also increase the computation time. 
%
Logistic Regression is probably the most basic classifier that still provides reasonably good results for a wide variety of problems. It uses a logistic function to determine the probability of a value belonging to a class or not. \emph{C} parameter represents the inverse of the regularization term and is important to prevent overfitting. We also explored the \emph{class\us weight} parameter which sets weights to sample classes inversely proportional to class frequencies in the input data (for \emph{balanced}). If \emph{None} is given, all classes have the same weight. Finally, \emph{penalty} parameter specifies the norm to use when computing the cost function.  
%
To have an idea about the performance of small and shallow neural networks on small datasets, we tried Multilayer Perceptron (MLP) classifier. It comes with a rich set of parameters such as \emph{alpha} which is the regularization term, \emph{solver} which is the weight optimization algorithm used during training or \emph{activation} that is the function used in each neuron of hidden layers to determine its output value. The most critical parameter is \emph{layer\us sizes} that specifies the number of neurons in each hidden layer. We tried many tuples such as (10, 1), (20, 1), $\dots $, (100, 4) where the first number is for the neurons and the second for the layer they belong to.   
%
Same as Logistic Regression, Na\"ive Bayes is also a very simple and popular classifier that provides high-quality solutions to many problems. It is based on Bayes theorem: 
\begin{equation}
P(A \mid B) = \frac{P(B \mid A) \, P(A)}{P(B)}
\end{equation}
which shows a way to get the probability of A given evidence B. For Na\"ive Bayes, we probed \emph{alpha} which is the smoothing parameter (dealing with words not in training data) and \emph{fit\us prior} for learning (or not) class prior probabilities.
%
%
The last algorithm we explored is Maximum Entropy classifier. It is a generalization of Na\"ive Bayes providing the possibility to use a single parameter for associating a feature with more than one label and captures the frequencies of individual joint-features. We explored four of the implementation methods that are available.
All algorithms, their parameters, and the grid-searched values are summarized in Table~\ref{table:algpars}.    
\section{RESULTS}
\label{sec:results}
%
\begin{table*}[!t]
\centering
\newcolumntype{C}[1]{>{\centering\arraybackslash} m{#1}}
\begin{tabular}{ | C{1.5cm} C{0.9cm} C{10.1cm} | }
\hline
\vspace{1mm}
\bf Algorithm & \bf Step & \bf Optimal Parameter Values \\ 
\hline
\multirow{2}{*}{SVM} 
& vect & ngram\us range: (1, 2), smooth\us idf: False, stop\us words: None, norm: l2 \\
& clf & C: 100, gamma: 0.005, kernel: rbf  \\ 
\hline
\multirow{2}{*}{NuSVM} 
 & vect & ngram\us range: (1, 2), smooth\us idf: True, stop\us words: None, norm: l2  \\ 
 & clf & kernel: linear, nu: 0.5 \\
\hline
\multirow{2}{*}{RF} 
& vect & ngram\us range: (1, 1), smooth\us idf: False, stop\us words: None, norm: l2   \\ 
& clf & max\us depth: 90, max\us feat: sqrt, n\us est: 700   \\
\hline
\multirow{2}{*}{LR} 
& vect & ngram\us range: (1, 2), smooth\us idf: False, stop\us words: None, norm: None   \\ 
& clf & C: 0.01, class\us weight: balanced, penalty: l2  \\
\hline
\multirow{2}{*}{MLP} 
& vect & ngram\us range: (1, 1), smooth\us idf: True, stop\us words: None, norm: l2   \\
& clf & alpha: 0.05, layer\us sizes: (60, 2), activation: relu, solver: adam  \\
\hline
\multirow{2}{*}{NB} 
& vect & ngram\us range: (1, 2), smooth\us idf: False, stop\us words: None, norm: l2   \\
& clf & alpha: 0.1, fit\us prior: True  \\ 
\hline
\multirow{2}{*}{ME} 
& vect & ngram\us range: (1, 1), smooth\us idf: False, stop\us words: None, norm: l2   \\
& clf & method: iis \\
\hline
\end{tabular}
\caption{Best parameter values and scores for Facebook data}
\label{table:bestfb}
\end{table*}
\subsection{Optimal Parameter Values}
\label{ssec:optvalues}
%
We performed 5-fold cross-validation grid-searching in the train part of each dataset (90\,\% of samples). The best parameters of the vectorization and classification step for each algorithm on the Facebook data are presented in Table~\ref{table:bestfb}. The corresponding results on the Mall data are presented in Table~\ref{table:bestmall}. Regarding \emph{Tf-Idf} vectorizer, we see that adding bigrams is fruitful in most of the cases (9 out of 14). Smoothing \emph{Idf} on the other hand does not seem necessary. Regarding stop words, keeping every word (stop\us words=None) gives the best results in 13 from 14 cases. Removing Czech stop words gives the best score with Random Forest on Mall data only. As for normalization, using \emph{l2} seems the best practice in most of the cases (10 out of 14).
%
\begin{table*}[!t]
\centering
\newcolumntype{C}[1]{>{\centering\arraybackslash} m{#1}}
\begin{tabular}{ | C{1.5cm} C{0.9cm} C{10.1cm} | }
\hline
\vspace{1mm}
\bf Algorithm & \bf Step & \bf Optimal Parameter Values \\ 
\hline
\multirow{2}{*}{SVM} 
& vect & ngram\us range: (1, 2), smooth\us idf: False, stop\us words: None, norm: l2 \\
& clf & C: 100, gamma: 0.01, kernel: rbf  \\ 
\hline
\multirow{2}{*}{NuSVM} 
& vect & ngram\us range: (1, 2), smooth\us idf: False, stop\us words: None, norm: l2  \\ 
& clf & kernel: linear, nu: 0.45 \\
\hline
\multirow{2}{*}{RF} 
& vect & ngram\us range: (1, 1), smooth\us idf: True, stop\us words: Czech, norm: None   \\ 
& clf & max\us depth: 90, max\us feat: sqrt, n\us est: 100   \\
\hline
\multirow{2}{*}{LR} 
& vect & ngram\us range: (1, 2), smooth\us idf: True, stop\us words: None, norm: l2   \\ 
& clf & C: 10, class\us weight: None, penalty: l2  \\
\hline
\multirow{2}{*}{MLP} 
& vect & ngram\us range: (1, 2), smooth\us idf: False, stop\us words: None, norm: l2   \\
& clf & alpha: 0.01, layer\us sizes: (40, 2), activation: relu, solver: adam \\
\hline
\multirow{2}{*}{NB} 
& vect & ngram\us range: (1, 2), smooth\us idf: True, stop\us words: None, norm: l1   \\
& clf & alpha: 0.05, fit\us prior: False  \\ 
\hline
\multirow{2}{*}{ME} 
& vect & ngram\us range: (1, 1), smooth\us idf: False, stop\us words: None, norm: l1   \\
& clf & algorithm: iis \\
\hline
\end{tabular}
\caption{Best parameter values and scores for Mall data}
\label{table:bestmall}
\end{table*}
%
Regarding classifier parameters, we see that SVM performs better with \emph{rbf} kernel and C = 100 in both datasets. The \emph{linear} kernel is the best option for NuSCV instead. In the case of Random Forest, we see that a max\us depth of 90 (the highest we tried) and \emph{sqrt} \emph{max\us feat} are the best options. Higher values could be even better. In the case of Logistic Regression, the only parameter that showed consistency on both dataset is \emph{penalty} (l2). We also see that MLP is better trained with \emph{relu} activation function and \emph{adam} optimizer. Finally, the two parameters of Na\"ive Bayes did not show any consistency on the two datasets whereas \emph{iis} was the best methods for Maximum Entropy in both of them.
%
\begin{table*}[!t]
	\centering      
	\setlength\tabcolsep{7pt} 
	\newcolumntype{C}[1]{>{\centering\arraybackslash} m{#1}} 
	\begin{tabular}
		{| c | c c c | c c c | }
		\hline
		& \multicolumn{6}{ c |}{\textbf{Facebook} \qquad \qquad \qquad \qquad \qquad
			\textbf{Mall}~~~} \\ [0.12ex] 
		\textbf{Algorithm} & \bf GS Acc & \bf Test Acc & 
		\bf Test F$_1$ & \bf GS Acc & \bf Test Acc & \bf Test F$_1$ \\ [0.12ex] 
		\hline
		SVM & \underline{70.4} & \underline{69.7} & \underline{63.2} & {\bf 93.1} & {\bf 92.1} & {\bf 91.6}  \\ [0.07ex]
		NuSVM & 69.8 & 69.3 & {\bf 64.9} & 92.6 & 91.9 & 91.4  \\ [0.07ex]
		RF & 65.8 & 62.7 & 44.2 & 88.3 & 85.5 & 83.8 \\ [0.07ex]
		LR & {\bf 70.8} & {\bf 69.9} & 62.9 & \underline{92.8} & 91.8 &  91.3  \\ [0.07ex]
		MLP & 66.5 & 64.1 & 59.2 & 90.1 & 89.8 & 86.4  \\ [0.07ex]
		NB & 68.7 & 67.2 & 57.6 & 92.8 & \underline{92} & \underline{91.5}  \\ [0.07ex]
		ME & 67.9 & 66.8 & 57.5 & 91.6 & 91.9 & 90.7 \\ [0.07ex]
		\hline 
	\end{tabular}
	\caption{Top grid-search and test scores for each algorithm}  
	\label{table:topscores}
\end{table*}
%

\begin{table*}[!t]
	\centering      
	\setlength\tabcolsep{7pt} 
	\newcolumntype{C}[1]{>{\centering\arraybackslash} m{#1}} 
	\begin{tabular}
		{| c | c c | c c | }
		\hline
		& \multicolumn{4}{ c |}{\textbf{Facebook} \qquad \qquad \qquad
			\textbf{Mall}~~~} \\ [0.12ex] 
		\textbf{Algorithm} & \bf Test Acc & 
		\bf Test F$_1$ & \bf Test Acc & \bf Test F$_1$ \\ [0.12ex] 
		\hline
		\textsc{SVM} & 69.1 & 62.7 & 91.9 & 90.4 \\ [0.07ex]
		\textsc{LR} & 69.8 & 63.1 & 92.2 & 91.6  \\ [0.07ex]
		\textsc{NB} & 65.7 & 57.4 & 91.8 & 91.4   \\ [0.07ex]
		\hline 
	\end{tabular}
	\caption{AdaBoost scores for the top three algorithms}  
	\label{table:adascores}
\end{table*}
\subsection{Test Scores Results}
\label{ssec:testscores}
%
We used the best performing vectorizer and classifier parameters to assess the classification performance of each algorithm in both datasets. The top grid-search accuracy, test accuracy and test macro F$_1$ scores are shown in Table~\ref{table:topscores}. For lower variance, the average of five measures is reported. The top scores on the two datasets differ a lot. That is because Facebook data classification is a multiclass discrimination problem (\emph{negative} vs. \emph{neutral} vs. \emph{positive}), in contrast with Mall review analysis which is purely binary (\emph{negative} vs. \emph{positive}). As we can see, Logistic Regression and SVM are the top performers in Facebook data. NuSVM and Na\"ive Bayes perform slightly worse. MLP and Random Forest, on the other hand, fall discretely behind. On the Mall dataset, SVM is dominant in both accuracy and F$_1$. It is followed by Logistic Regression, Na\"ive Bayes and NuSVM. Maximum Entropy is near whereas MLP and Random Forest are again considerably weaker. Similar results are also reported in other works like \cite{sheshasaayee2017comparison} where again, SVM and Na\"ive Bayes outrun Random Forest on text analysis tasks. From the top three algorithms, Na\"ive Bayes was the fastest to train, followed by Logistic Regression. SVM was instead considerably slower. The 91.6\,\% of SVM in F$_1$ score on Mall dataset is considerably higher than the 78.1\,\% F$_1$ score reported in Table~7 of \cite{DBLP:conf/konvens/VeselovskaHS12}. They used Na\"ive Bayes with $ \alpha = 0.005 $ and 5-fold cross-validation, same as we did here. Unfortunately, no other direct comparisons with similar studies are possible.   
\subsection{Boosting Results}
\label{ssec:boosting}
%
We picked SVM, Logistic Regression and Na\"ive Bayes with their corresponding optimal set of parameters and tried to increase their performance further using Adaptive Boosting \cite{FREUND1997119}. AdaBoost is one of the popular mechanisms for reinforcing the prediction capabilities of other algorithms by combining them in a weighted way. It tries to tweak future classifiers based on the wrong predictions of the previous ones and selects only the features known to improve prediction quality. On the negative side, AdaBoost is sensitive to noisy data and outliers which means that it requires careful data preprocessing. First, we experimented with a few estimators in Adaboost and got poor results in both datasets. Increasing the number of estimators increased accuracy and F$_1$ scores until some point (about 5000 estimators) and was further useless. The detailed results are presented in Table~\ref{table:adascores}. As we can see, no improvements over the top scores of each algorithm were gained. The results we got are actually slightly lower.  
%
As a final attempt, we combined SVM, LR, and NB in a majority voting ensemble scheme. Test accuracy and F$_1$ scores on Facebook were 69.5 and 64.2\,\%, respectively. The corresponding results on Mall dataset were 92.1 and 91.5\,\%. Again, we see that the results are slightly lower than top scores of the tree algorithms and no improvement was gained on either of the datasets.     
\section{CONCLUSIONS}
\label{sec:disc}
%
In this paper, we tried various supervised learning algorithms for sentiment analysis of Czech texts using two existing datasets of Facebook posts and Mall product reviews. We grid-searched various parameters of \emph{Tf-Idf} vectorizer and each of the machine learning algorithms. According to our observations, best sentiment predictions are achieved when bigrams are added, Czech stop words are not removed and \emph{l2} normalization is applied in vectorization. We also reported the optimal parameter values of each explored classifier which can serve as guidelines for other researchers. The accuracy and F$_1$ scores on the test part of each dataset indicate that the best-performing algorithms are Support Vector Machine, Logistic Regression, and Na\"ive Bayes. Their simplicity and speed make them optimal choices for sentiment analysis of texts in cases when few thousands of sentiment-labeled data samples are available.    
%
We also observed that ensemble methods like bagging (e.g., Random Forest), boosting (e.g., AdaBoost) or even voting ensemble schemes do not add value to any of the three basic classifiers. This is probably because all \emph{Tf-Idf} vectorized features are relevant and necessary for the classification process and extra combinations of their subsets are not able to further improve classification performance.
%
Finally, as future work, we would like to create additional labeled datasets with texts of other languages and perform similar sentiment analysis experiments. That way, valuable metalinguistic insights could be drawn and reported.    
\section*{ACKNOWLEDGEMENTS}
\noindent The research was [partially] supported by OP RDE project No. CZ.02.2.69/0.0/0.0/16\_027/0008495, International Mobility of Researchers at Charles University.
%
\bibliographystyle{apalike}
{\small
\bibliography{saczech}}

%
%

\vfill
\end{document}